\documentclass[letterpaper, 10 pt, conference]{ieeeconf}  % Comment this line out if you need a4paper

\IEEEoverridecommandlockouts                              % This command is only needed if 
\usepackage{amsmath}
\usepackage{mathtools}

\usepackage{amsthm}
\usepackage{amsfonts}
\usepackage{xcolor}
\usepackage{graphicx}
\usepackage{booktabs} % for professional tables
\newcommand{\nonl}{\renewcommand{\nl}{\let\nl\oldnl}}
\usepackage{multirow,array}
\usepackage{subcaption}
\usepackage[capitalize,noabbrev]{cleveref}
\usepackage[ruled,linesnumbered]{algorithm2e}
\usepackage{algpseudocode}

\crefname{algocf}{alg.}{algs.}
\Crefname{algocf}{Algorithm}{Algorithms}

\theoremstyle{plain}

\theoremstyle{definition}

\theoremstyle{remark}

\title{\LARGE \bf
Diffusion Policies for Risk-Averse Behavior Modeling in Offline Reinforcement Learning}

\author{Xiaocong Chen$^{1}$, Siyu Wang$^{2}$, Tong Yu$^{3}$ and Lina Yao$^{1,4}$% <-this % stops a space
\thanks{$^{1}$Xiaocong Chen and Lina Yao are with Data 61, CSIRO, Australia
        {\tt\small xiaocong.chen@data61.csiro.au}}%
\thanks{$^{2}$Siyu Wang is with School of Computing, Macquarie University, Australia; $^{3}$ Tong Yu is with Adobe Research, USA;$^{4}$ Lina Yao is also with the School of Computer Science and Engineering, University of New South Wales, Australia}
}

\begin{document}

\maketitle
\thispagestyle{empty}
\pagestyle{empty}

%%%%%%%%%%%%%%%%%%%%%%%%%%%%%%%%%%%%%%%%%%%%%%%%%%%%%%%%%%%%%%%%%%%%%%%%%%%%%%%%
\begin{abstract}

Offline reinforcement learning (RL) presents distinct challenges as it relies solely on observational data. A central concern in this context is ensuring the safety of the learned policy by quantifying uncertainties associated with various actions and environmental stochasticity. Traditional approaches primarily emphasize mitigating epistemic uncertainty by learning risk-averse policies, often overlooking environmental stochasticity.
In this study, we propose an uncertainty-aware distributional offline RL method to simultaneously address both epistemic uncertainty and environmental stochasticity. We propose a model-free offline RL algorithm capable of learning risk-averse policies and characterizing the entire distribution of discounted cumulative rewards, as opposed to merely maximizing the expected value of accumulated discounted returns.
Our method is rigorously evaluated through comprehensive experiments in both risk-sensitive and risk-neutral benchmarks, demonstrating its superior performance.
\end{abstract}

%%%%%%%%%%%%%%%%%%%%%%%%%%%%%%%%%%%%%%%%%%%%%%%%%%%%%%%%%%%%%%%%%%%%%%%%%%%%%%%%
\section{INTRODUCTION}

In safety-critical applications, deploying online reinforcement learning (RL) is often infeasible due to the extensive exploration required, which can result in hazardous or costly outcomes. Offline RL~\cite{levine2020offline} addresses this issue by enabling the learning of optimal policies from pre-collected data, thereby mitigating the risks associated with direct exploration. However, safety-critical environments require policies that not only perform well but also manage risk effectively. Risk in these settings arises from the unpredictability of the environment's behavior or the generalization ability of the learned policy in unseen situations, making risk-aware decision-making essential.

In most existing offline RL approaches, the primary focus is on maximizing expected returns~\cite{levine2020offline}, often without adequately accounting for the variability and potential adverse consequences of certain actions. This risk-neutral perspective can result in suboptimal performance, particularly in scenarios where avoiding worst-case outcomes is critical. Uncertainty in the quality and consistency of the collected data in offline RL is closely associated with risk, as unreliable or inconsistent data can lead to the development of policies that perform poorly in practice.

To address these risks, current risk-aware methods, such as those based on conditional variational autoencoders (cVAEs), attempt to mitigate bootstrapping errors by reconstructing behavior policies~\cite{urp2021riskaverse,lyu2022mildly}. However, these methods frequently rely on imitation learning, which aligns the learned policy with the observed behavior in the dataset. This dependence makes them susceptible to suboptimal performance when the dataset contains diverse or suboptimal trajectories~\cite{ma2021conservative}. Furthermore, cVAEs are limited in their ability to model complex behavior distributions, a limitation that becomes particularly problematic in offline RL settings where heterogeneous datasets—combining trajectories from multiple behavior policies—are common. These heterogeneous datasets present significant challenges in developing policies that are both high-quality and risk-aware~\cite{pearce2023imitating,wang2023diffusion}. While one such method attempts to refine behavior policies~\cite{ma2021conservative}, it often lacks the flexibility required to generalize effectively across different environments and tasks, further complicating risk management.

To address the aforementioned challenges, it is essential to account for uncertainty when managing risk, as it reflects both the unpredictability of the environment and the quality of the pre-collected data. In response, we propose Uncertainty-aware offline Distributional Actor-Critic (UDAC), a novel approach that integrates uncertainty-aware policy learning with risk management. UDAC leverages diffusion models, which have demonstrated strong capabilities in modeling behavior policies~\cite{pearce2023imitating}, to address both risk and uncertainty within a unified framework.

Our method introduces several key innovations: (i) \textit{Elimination of Manual Behavior Policy Specification}: UDAC automates the modeling of behavior policies, removing the need for manual design and making it adaptable to a wide variety of risk-sensitive tasks. (ii) \textit{Accurate Behavior Policy Modeling}: By utilizing a controllable diffusion model, UDAC accurately captures the full distribution of behavior policies, which is critical for robust decision-making in environments with substantial uncertainty and risk. iii) UDAC incorporates a \textit{perturbation model tailored for risk-sensitive settings}, allowing it to balance exploration and caution. This model introduces controlled variability into the policy, enabling UDAC to manage risk more effectively by accounting for uncertainties in both the environment and the dataset.

Extensive experimental evaluations on benchmarks, including risk-sensitive D4RL, risky robot navigation, and risk-neutral D4RL, demonstrate that UDAC surpasses baseline methods in risk-sensitive tasks while maintaining competitive performance in risk-neutral settings. These results indicate that UDAC offers a comprehensive framework for managing both risk and uncertainty in offline RL, positioning it as an effective tool for navigating complex, safety-critical environments.

\section{Related Work} 
% diffusion
\cite{pearce2023imitating} suggest that the imitation of human behaviors can be improved through the use of expressive and stable diffusion models.~\cite{janner2022planning} employ a diffusion model as a trajectory generator in their Diffuser approach, where a full trajectory of state-action pairs forms a single sample for the diffusion model.
% A separate return model is trained to predict the cumulative rewards of each trajectory sample, and its guidance is integrated into the reverse sampling stage. 
% This methodology is comparable to Decision Transformer~\cite{chen2021decision}, which employs GPT2 to learn a trajectory generator with the assistance of true trajectory returns. However, online use of sequence models becomes problematic as they are unable to predict actions from states autoregressively, since the states are the result of the environment. As a result, during evaluation, an entire trajectory is predicted for each state, with only the first action being applied, resulting in a significant computational expense. To address this,~\cite{wang2023diffusion} suggests using diffusion to learn the policy via policy improvement. 
% distributional
~\cite{singh2020improving} utilize a distributional critic in their algorithm, but it is restricted to the CVaR and limited to the online reinforcement learning (RL) setting. Furthermore, their use of a sample-based distributional critic renders the computation of the CVaR inefficient. To enhance the computational efficiency of the CVaR,~\cite{urp2021riskaverse} extend the algorithm to the offline setting.~\cite{ma2021conservative} replace normal Q-learning with conservation offline Q-learning and an additional constant to penalize out-of-distribution (OOD) behaviors. However,~\cite{lyu2022mildly} observe that the introduced extra constant is excessively severe and propose a more gentle penalized mechanism that uses a piecewise function to control when applying the extra penalization.~\cite{queeney2024risk} focuses on the safe offline RL domain which also uses the risk-averse approach.

% \textbf{Differences between existing works.} There are three similar works that have similar
\noindent\textbf{Differences between existing works.} There are two similar works that also apply the diffusion model in offline reinforcement learning: Diffusion-QL~\cite{wang2023diffusion} and Diffuser~\cite{janner2022planning}. Diffuser~\cite{janner2022planning} is from the model-based trajectory-planning perspective, while Diffusion-QL~\cite{wang2023diffusion} is from the offline model-free policy-optimization perspective. However, our work is from the risk-sensitive perspective and extends it into the distributional offline reinforcement learning settings.

\section{Background}
In this work, a Markov Decision Process (MDP) is considered with states $s \in \mathcal{S}$ and actions $a \in \mathcal{A}$ that may be continuous. It includes a transition probability $P(\cdot|s,a):\mathcal{S}\times\mathcal{A}\times\mathcal{S}\rightarrow [0,1]$, reward function $R(\cdot|s,a):\mathcal{S}\times\mathcal{A}\rightarrow\mathbb{R}$, and a discount factor $\gamma\in[0,1)$. A policy $\pi(\cdot|s)$ is defined as a mapping from states to the distribution of actions. RL aims to learn $\pi(\cdot|s)$ such that the expected cumulative long-term rewards $\mathbb{E}_{s_0,a_t\sim\pi(\cdot|s_t),s_{t+1}\sim P(\cdot|s_t,a_t)}[\sum_{t=0}^\infty \gamma^t R(\cdot|s_t,a_t)]$ are maximized. The
state-action function $Q_\pi(s,a)$ measures the discounted return starting from state $s$ and action $a$ under policy $\pi$. We assume the reward function $R(\cdot|s,a)$ is bounded, i.e., $|R(\cdot|s,a)|\leq r_{max}$. Now, consider the policy $\pi(\cdot|s)$, the Bellman operator $\mathcal{T}^\pi:\mathbb{R}^{|\mathcal{S}||\mathcal{A}|}\rightarrow\mathbb{R}^{|\mathcal{S}||\mathcal{A}|}$ is introduced by~\cite{riedmiller2005neural} to update the corresponding $Q$ value,
\begin{align}
    \mathcal{T}^\pi Q(s,a)\mathrel{:=} \mathbb{E}[R(s,a)]+\gamma\mathbb{E}_{P(s'|s,a)\pi(a'|s')}[Q(s',a')].
\end{align}

In offline RL, the agent can only access a fixed dataset $\mathcal{D}\mathrel{:=}\{(s,a,r,s')\}$, where $r\sim R(\cdot|s,a), s'\sim P(\cdot|s,a)$. $\mathcal{D}$ comes from the behavior policies. We use $\pi_b$ to represent the behavior policy, and $(s,a,r,s')\sim\mathcal{D}$ represents the uniformly random sampling. 

In distributional RL~\cite{bellemare2017distributional}, the aim is to learn the distribution of discounted cumulative rewards, represented by the random variable $Z^\pi(s,a) = \sum_{t=0}^\infty \gamma^t R(\cdot | s_t, a_t)$. Similar to the $Q$ function and Bellman operator mentioned earlier, we can define a distributional Bellman operator as follows,
\begin{align}
    \mathcal{T}^\pi Z(s,a)\mathrel{:=}_D R(s,a)+\gamma Z (s',a').\notag \\ 
    \quad s'\sim P(\cdot|s,a), a'\in \pi(\cdot|s')\label{eq:dbo}.
\end{align}
The symbol $\mathrel{:=}_D$ is used to signify equality in distribution. Thus, $Z^\pi$ can be obtained by applying the distributional Bellman operator $\mathcal{T}^\pi$ iteratively to the initial distribution $\mathcal{Z}$. In offline settings, $\mathcal{T}^\pi$ can be approximated by $\hat{\mathcal{T}}^\pi$ using $\mathcal{D}$. Then, we can compute $Z^\pi$ by starting from an arbitrary $\hat{Z}^0$, and iteratively minimizing the p-Wasserstein distance between $Z$ and $\hat{\mathcal{T}}^\pi\hat{Z}^\pi$ (To simplify, we will use the $W_p(\cdot)$ to represent the p-Wasserstein distance). And the measure metric $\overline{d}_p$ over value distributions is defined as, 
\begin{align*}
    \overline{d_p} (Z_1,Z_2) = \sup_{s,a}W_p(Z_1(s,a),Z_2(s,a)), Z_1,Z_2\in\mathcal{Z}.
\end{align*}

% \begin{theorem}(Introduced by~\cite{bellemare2017distributional})
% The operator $\mathcal{T}^\pi$ is a $\gamma$-contraction in $\overline{d_p}$, which can converge to the optimal action-state value $Q^*$ by taking expectation $\mathbb{E}[Z]$. \label{theo1}
% \end{theorem}

\section{Methodology}
% Our proposed method, UDAC, is based on the quantile-based distributional RL framework, comprising three parts: 
% quantile value network, policy network $\pi_{\theta}(s)$ and the behavior policy network $\pi_{db}$.
% The overall framework for UDAC is outlined in~\Cref{sec:critic}. 
% The loss of the actor is defined as the risk-distortion operator\footnote{risk-distortion operator is a class of distortion function that also considers the risk~\cite {wang2000class}.} applied to the learned return distribution and optimized through gradient-based methods. With this actor-critic setup, a risk-averse criterion can be optimized.
% However, in the offline setting, controlling the bootstrapping error is crucial. To address this, a diffusion policy is incorporated in the actor to learn a generative model of the behavior policy, which is described in~\Cref{sec:diffusion}. Finally, all components are combined in~\Cref{sec:overall} to implement the algorithm for a given risk distortion $\mathbb{D}$.
\subsection{Overview of UDAC}\label{sec:critic}
In the critic network of UDAC, our aim is policy evaluation, and we rely on \Cref{eq:dbo} to achieve this objective.
Following the approaches of \cite{dabney2018implicit,urp2021riskaverse,ma2021conservative}, we implicitly represent the return through its quantile function, as demonstrated by \cite{dabney2018implicit}.
In order to optimize the $Z$, we utilise a neural network with a learnable parameter $\theta$ to represent the quantile function. We express such implicit quantile function as $F_Z^{-1}(s,a;\tau)$ for $Z$, where $\tau\in[0,1]$ is the quantile level. Notice that, the original quantile function can only be used for discrete action. We can extend it into a continuous setting by considering all $s, a$, and $\tau$ as the inputs and only the quantile value as the output. To learn the $\theta$, the common strategy is the fitted distributional evaluation using a quantile Huber-loss $\mathcal{L}_{\mathcal{K}}$~\cite{huber1992robust}.
Our optimizing objective for the critic network is,
\begin{align}
    \mathcal{L}(\hat{\mathcal{T}}^\pi \hat{Z}^\pi,Z) = W_p(\hat{\mathcal{T}}^\pi \hat{Z}^\pi,Z)\simeq \mathcal{L}_{\mathcal{K}}(\delta;\tau), \notag \\ \text{ with } \delta=r+\gamma F_{Z'}^{-1}(s',a';\tau') - F_Z^{-1}(s,a;\tau), \label{eq:4}
\end{align}
where $(s,a,r,s')\sim\mathcal{D}$ and $a'\sim\pi(\cdot|s')$. The $\delta$ is also known as the distributional TD error~\cite{dabney2018distributional}. And the quantile Huber-loss $\mathcal{L}_{\mathcal{K}}$ is represented as,
\begin{align}
    \mathcal{L}_{\mathcal{K}}(\delta;\tau) = 
    \begin{cases} |1_{\delta<0}-\tau| \cdot \delta^2 / (2\mathcal{K}) & \text{if } |\delta|<\mathcal{K},\\ 
    |1_{\delta<0}-\tau|\cdot (\delta - \mathcal{K}/2) & \text{otherwise}
    \end{cases} \label{eq:5}
\end{align}
where $\mathcal{K}$ is the threshold. Notice that there is a hyper-parameter $\tau$ which may affect our objective function.
Here, we use a simple strategy to control the $\tau$: Uniformly random sampling $\tau\sim U(0,1)$. It is worth mentioning that there is an alternative way for controlling $\tau$, which is a quantile level proposal network $P_\psi(s, a)$~\cite{yang2019fully}. To simplify the overall model and reduce the number of hyperparameters, we just use the simplified random sampling approach and leave the quantile proposal network for future work. The overall loss function for the critic is
\begin{align}
     \mathbb{E}_{(s,a,r,s')\sim \mathcal{D}, a'\sim \pi(\cdot|s')}\bigg[\frac{1}{N*K}\sum_{i=0}^N\sum_{j=0}^K\mathcal{L}_{\mathcal{K}}(\delta_{\tau_i,\tau'_j};\tau_i)\bigg],    \label{eq:criticloss}
\end{align}
where $N, K$ are the numbers of quantile levels.
In the risk-sensitive setting, deterministic policies are favoured over stochastic policies due to the latter's potential to introduce additional randomness, as noted by~\cite{pratt1978risk}. Moreover, in offline settings, there is no benefit to exploration commonly associated with stochastic policies. Therefore, the focus is on parameterized deterministic policies, represented by $\pi_\phi(s):\mathcal{S}\rightarrow\mathcal{A}$. Given the risk-distortion function $\mathbb{D}$, the loss function of actor is written as,
\begin{align}
-\mathbb{E}_{s\sim \rho_b}[\mathbb{D}(F_Z^{-1}(s,\pi_{\phi}(s);\tau))] \label{eq:loss_distortion}
\end{align}
with the marginal state distribution $\rho_b$ from the behavior policy. To optimize it, we can backpropagate it through the learned critic at the state that sampled from the behavior policy. It is equivalent to maximising risk-averse performance. Then, combining the above process with the SAC~\cite{haarnoja2018soft} leads to our novel UDAC model.

\subsection{Generalized Behavior Policy Modeling}\label{sec:diffusion}
% UDAC relies on behaivor policy modeling to learn the final policy.
We separate the actor into two components: an imitation component~\cite{fujimoto2019off} and a conditionally deterministic perturbation model $\xi$, 
% \textcolor{red}{replace with a OTP perturbation?}
\begin{align}
    \pi_\phi(s) = \lambda\xi(\cdot|s,\beta) + \beta, \beta\sim\pi_b, \label{eq:actoroverall}
\end{align}
where $\xi(\cdot|s,\beta)$ can be optimized by using the actor loss function(i.e., $-\mathbb{E}_{s\sim \rho_b}[\mathbb{D}(F_Z^{-1}(s, a;\tau))]$) that was defined previously, and $\beta$ is the action that sampled from behavior policy $\pi_b$, $\lambda$ is a scale that used to control the magnitude of the perturbation. As aforementioned, we prefer the deterministic policy over the stochastic policy because of the randomness. However, the bootstrapping error will appear on the offline RL setting. Existing works~\cite{urp2021riskaverse,wu2019behavior,ma2021conservative,lyu2022mildly} use policy regularization to regularize how far the policy can deviate from the behavior policy. But the policy regularization will adversely affect the policy improvement by limiting the exploration space, and only suboptimal actions will be observed. Moreover, the fixed amount of data from the offline dataset may not be generated by a single, high-quality behaviour policy and thus affect the imitation learning performance (i.e., the logged data may come from different behavior policies). Hence, behavioral cloning~\cite{pomerleau1991efficient} would not be the best choice. Moreover, behavioral cloning suffers from mode-collapse, which will affect actor's optimization.  Here, we attempt to use the generative method to model the behaviour policy from the observed dataset. The most popular direction is the cVAE-based approaches~\cite{urp2021riskaverse,wu2019behavior} which leverages the capability of the auto-encoder to reconstruct the behavior policy. 
% However the behavior policies on the offline dataset are normally collected from different agents and may have different modalities. It implies that some behavior policies may be sub-optimal. In this work, we treat these sub-optimal policies as noisy policies. Inspired by recent work of denoising in computer vision related tasks~\cite{ho2020denoising}, we employ the diffusion model instead of cVAE to model the behavior policy. 
However, these methods did not account for the noisy behavior policies resulting from the offline reinforcement learning environment. This environment typically comprises datasets collected from various agents, each potentially exhibiting distinct modalities. This suggests that certain behavior policies within these datasets might be sub-optimal, adversely affecting overall performance. 
To address this, we seek denoising diffusion probabilistic model (DDPM)~\cite{ho2020denoising} to facilitate robust behavior policy modelling.
% The diffusion model naturally admits a progressive lossy decompression scheme that can be interpreted as a generalization of autoregressive decoding. Hence, by employing such a behavior policy model, UDAC will leverage those pre-collected sub-optimal trajectories to learn the final policy.
The diffusion model facilitates a gradual lossy decompression approach, which extends autoregressive decoding. Thus, UDAC utilizes this model as a behavior policy to utilize pre-gathered sub-optimal trajectories for final policy learning.

% By employing such a behavior policy model, those pre-collected sub-optimal data will be treated as ``noisy optimal''.  
% Different from existing works that use cVAE, we use the diffusion model instead~\cite{ho2020denoising}.

We represent the behavior policy by using the reverse process of a conditional diffusion model, which is denoted by the parameter $\psi$, as follows,
\begin{align}
    \pi(a|s) = p(a^{0:N}|s)=\mathcal{N}(a^N;0,I)\sum_{i=1}^N p(a^{i-1}|a^i,s), \label{eq:piinbays}
\end{align}
where $p(a^{i-1}|a^i,s)$ can be represented as a Gaussian distribution $\mathcal{N}(a^{i-1};\mu_\psi(a^i,s,i),\sum_\psi(a^i,s,i))$, where $\sum_\psi(a^i,s,i)$ is a covariance matrix. To parameterize $p(a^{i-1}|a^i,s)$ as a noise prediction model, we fix the covariance matrix as $\sum_\psi(a^i,s,i)=\beta_i I$ and construct the mean as,
\begin{align}
    \mu_\psi(a^i,s,i) = \frac{1}{\sqrt{\alpha_i}}\Bigg(a^i-\frac{\beta_i}{\sqrt{1-\overline{\alpha}_i}}\epsilon_\psi(a^i,s,i)\Bigg).
\end{align}
To obtain a sample, we begin by sampling $a^N$ from a normal distribution with mean 0 and identity covariance matrix, $\mathcal{N}(0,I)$. We then sample from the reverse diffusion chain, which is parameterized by $\psi$, to complete the process,
\begin{align}
    a^{i-1}|a^i = \frac{a^i}{\sqrt{\alpha_i}}-\frac{\beta_i}{\alpha_i\sqrt{1-\overline{\alpha}_i}}\epsilon_\psi(a^i,s,i) + \sqrt{\beta_i}\epsilon,\notag\\  \epsilon\sim\mathcal{N}(0,I), \text{ for } i = N,\cdots,1. \label{eq:acondition}
\end{align}
Following the approach of \cite{ho2020denoising}, we set $\epsilon$ to 0 when $i=1$ to improve the sampling quality. The objective function $\mathcal{L}_d(\psi)$ used to train the conditional $\epsilon$-model is,
\begin{align}
    \mathbb{E}_{i\sim \mathcal{U},\epsilon\sim\mathcal{N}(0,I),(s,a)\sim\mathcal{D}}\big[\|\epsilon - \epsilon_\psi(\sqrt{\overline{\alpha}_i}a+\sqrt{1-\overline{\alpha}_i}\epsilon,s,i)\|^2\big], \label{eq:diffusionloss} 
\end{align}
where $\mathcal{U}$ is a uniform distribution over the discrete set, denoted as $\{1,\cdots,N\}$. This diffusion model loss $\mathcal{L}_d(\psi)$ aims to learn from the behavior policy $\pi_b$ instead of cloning it. $\mathcal{L}_d(\psi)$ can be optimized by sampling from single diffusion step $i$ for each data point. However, it will be affected significantly by $N$, which is known as the bottleneck of the diffusion~\cite{kong2021fast}. 
% While it is not this paper's main goal, we just set the $N$ to be small.
We add this term to the actor loss and construct the final loss function. We then replace the term $\pi_b$ in the~\Cref{eq:actoroverall} with $d_\psi(s, a)$ (i.e., the diffusion behavior policy representation). 
% The empirical idea is using the generative method to model the behavior policy such as conditional Variational Auto-Encoder (cVAE)~\cite{urp2021riskaverse,lyu2022mildly} or just simply fix the behavior policy by using pre-defined one~\cite{ma2021conservative}.
% Existing methods modeling the behavior policy by using CVAE (ORAAC~\cite{urp2021riskaverse}, MCQ~\cite{lyu2022mildly}) or simply fixed (empirical- CODAC~\cite{ma2021conservative}). Here we uses diffusion instead.
% Why diffusion, as existing works can rarely learn high-quality behavioral policy (claimed as CODAC)

Nonetheless, as aforementioned, only a select number of trajectories are sub-optimal. We aim to control the diffusion process especially to denoise those sub-optimal trajectories. Specifically, we need to control the $a^{0:N}$ which is equivalent to decoding from the posterior~\Cref{eq:piinbays} (i.e., $p(a^{0:N}|s)$), and we can decompose this joint inference problem to a sequence of control problems at each diffusion step,
\begin{align}
    p(a^{i-1}|a^i,s)\propto p(a^{i-1}|a^i)\cdot p(s|a^{i-1},a^i).
\end{align}
We can further simplify $p(s|a^{i-1},a^i)=p(s|a^{i-1})$ via conditional independence assumptions from prior work on controlling diffusions~\cite{song2020score}. Consequently, for the $i$-step, we run gradient update on $a^{i-1}$,
\begin{align}
     \nabla_{a^{i-1}}&\log p(a^{i-1}|a^i,s) = \notag \\ & \nabla_{a^{i-1}}\log p(a^{i-1}|a^i) + \nabla_{a^{i-1}}\log p(s|a^{i-1}),
\end{align}
where both $\log p(a^{i-1}|a^i)$ and $\log p(s|a^{i-1})$ are differentiable: the first term is parametrized and can be calculated by~\Cref{eq:acondition}, and the second term is parametrized by a neural network classifier. Here, we have introduced a classifier to control the whole decoding process. We train the classifier on the diffusion latent variables and run gradient updates on the latent space $a^{i-1}$ to steer it towards fulfilling the control. These
image diffusion works take one gradient step towards $\nabla_{a^{i-1}}\log p(s|a^{i-1})$ per diffusion steps. The classifier is trained by using cross-entropy and can be easily added into the diffusion process (i.e., the overall loss function in~\Cref{eq:diffusionloss}.) 

% \subsection{Some Theoretical Analysis of the UDAC}
% We can observe that the distribution offline RL relies on the Bellman operator $\mathcal{T}^\pi$ and the behavior policy, we have the following important proposition of the UDAC.
% \begin{prop}Assume that $Q^\pi$ is the unique fixed point
% acquired by the $\mathcal{T}^\pi$ as per the definition of the Bellman operator, then in the support region of the behavior policy $\pi_b$, we denote the Q function of the behavior policy $\pi_b$ as $Q_{b}$ and the Q function of the learned optimal policy $Q_{b}^*$. We have $Q_b \leq Q^\pi \leq Q_b^*$. \label{prop1}
% \end{prop}
% \begin{prop}(Policy Improvement)
%     For all $(s,a) \in \mathcal{S}\times\mathcal{A}$, let $Q^\pi(s,a) = \mathbb{E}(Z^\pi(s,a))$, we have $Q^{\pi_{old}}(s,a) \leq Q^{\pi_{new}}(s,a)$ when minimizing the difference between the policy distribution and exponential form of soft action-value function. Where the $\pi$ is defined in~\Cref{eq:actoroverall}.\label{prop2}
% \end{prop} 
% \Cref{prop1} guarantees that the learned policy will perform at least as well as behavior policy, and~\Cref{prop2} shows that the soft policy improvement works under the distributional offline RL setting. The proof can be found on the~\Cref{appendix:proof}.

\subsection{Training Algorithm}\label{sec:overall}
By combing all of the components mentioned in previous sections, the overall training algorithm is described in~\Cref{alg:trainwithrandom}.
\begin{algorithm}[!h]
\SetAlgoLined
Initialize critic parameter $\theta$, actor parameter $\phi$, perturbation model $\xi$, quantile huber loss threshold $\mathcal{K}$, number of generated quantiles $N,K$, diffusion parameter $\psi$, distortion $\mathbb{D}$, learning rate $\eta$\; 
Sample quantile $\tau_i,\tau_j$, where $i=\{0,\cdots,N\}, j=\{0,\cdots,K\}$ from  $\mathcal{U}(0,1)$\;
\For{$t=1,\cdots$}{
    Sample $B$ transitions $(s,a,r,s')$ from the offline dataset\;
    Compute the $\delta$ based on~\Cref{eq:4}\;
    Sample $\beta$ from $d_\psi(s,a)$ and compute the policy $\pi_\phi$ based on~\Cref{eq:actoroverall}\;
    Compute critic loss by using~\Cref{eq:criticloss}\; 
    Compute the loss for $\xi$ by using~\Cref{eq:loss_distortion}\;
    Compute the diffusion loss by using~\Cref{eq:diffusionloss} \Comment{add the classifier loss we want to control the process} \;
    Compute the loss of actor by adding the loss of $\xi$ and the diffusion model\;
    Update $\theta$, $\phi$, $\psi$ accordingly\;
}
 \caption{UDAC with $U(0,1)$}\label{alg:trainwithrandom}
\end{algorithm}
% Note that, the update algorithm is based on SAC~\cite{haarnoja2018soft}, to make it simple, we just omit the target network initialization and update process.
The critic seeks to learn the return distribution, while the diffusion model aims to comprehend the behavioral policy, serving as a baseline action for the actor to implement risk-averse perturbations. As aforementioned, the distortion operator $\mathbb{D}$ is required. It can be any distorted expectation objective, such as CVaR, Wang, and CPW. In this work, we focus on $\text{CVaR}_{\alpha=0.1}$, which is the majority distorted measure used in recent literature~\cite{rigter2022one,urp2021riskaverse,ma2021conservative}.

%  Policy improvement
% \begin{lemma}
%  For all $(s,a)$
% \end{lemma}
\section{Experiments}\label{sec:exp}
In this section, we show that UDAC achieves state-of-the-art results on risk-sensitive offline RL tasks, including risky robot navigation and D4RL, and comparable performance with SOTAs in risk-neutral offline RL tasks. 

\subsection{Implementation Details}\label{appendix:implement}
In this section, we will describe the implementation details of the proposed UDAC.
\paragraph{Actor} As mentioned in the main text, the architecture of the actor is,
\begin{align*}
    \pi_\phi(s) = \lambda\xi(\cdot|s,\beta) + \beta, \beta\sim\pi_b,
\end{align*}
where $\xi: \mathcal{A}\rightarrow \mathbb{R}^{\|\mathcal{A}\|}$ (i.e., the traditional Q-learning component). And $\beta$ is the output of the behavioral model (i.e., the diffusion component).

If we want to integrate the proposed UDAC into an online setting, we can effortlessly remove the $\beta$ and set $\lambda = 1$. For our experiments, we construct the behavioral policy as an MLP-based conditional diffusion model, drawing inspiration from~\cite{ho2020denoising,nichol2021improved}. We employ a 3-layer MLP with Mish activations and 256 hidden units to model $\epsilon_\psi$. The input of $\epsilon_\psi$ is a concatenation of the last step’s action, the current state vector, and the sinusoidal positional embedding of timestep $i$. The output of $\epsilon_\psi$ is the predicted residual at diffusion timestep $i$. Moreover, we build two Q-networks with the same MLP setting as our diffusion policy, with 3-layer MLPs with Mish activations and 256 hidden units for all networks. Moreover, as mentioned in the main body, we set $N$ to be small to relieve the bottleneck caused by the diffusion. In practice, we use the noise schedule obtained under the variance preserving SDE~\cite{song2020score}.
\paragraph{Critic} For the critic architecture, it is built upon IQN~\cite{dabney2018implicit} but within continuous action space. We are employing a three-layer MLP with 64 hidden units.

\paragraph{Classifier} For the classifier which is used to control the diffusion process which is just a simple three-layer neural network with 256 hidden units. It was also pre-trained on the corresponding datasets.

We use Adam optimizer to optimize the proposed model. The learning rate is set to $0.001$ for the critic and the diffusion, $0.0001$ for the actor model. The target networks for the critic and the perturbation models are updated softly with $\mu= 0.005$.

\subsection{Experiments Setup}\label{appendix:expsetup}
The stochastic MuJoCo environments are set as follows with a stochastic reward modification. It was proposed by~\cite{urp2021riskaverse}, and the following description is adapted:
\begin{itemize}
    \item Half-Cheetah: $R_t(s,a) = \overline{r}_t(s,a) - 70 \mathcal{I}_{v>\overline{v}}\cdot\mathcal{B}_{0.1}$. where $\overline{r}_t(s,a)$ is the original environment reward, $v$ is the forward velocity, and $\overline{v}$ is a threshold velocity ($\overline{v}=4$ for Medium datasets and $\overline{v}=10$ for the Expert dataset). The maximum episode length is reduced to 200 steps.
    \item Walker2D/Hopper: $R_t(s,a) = \overline{r}_t(s,a) -p \mathcal{I}_{|\theta|>\overline{\theta}}\cdot\mathcal{B}_{0.1}$, where $\overline{r}_t(s,a)$ is the original environment reward, $\theta$ is the pitch angle, and $\overline{\theta}$ is a threshold velocity ($\overline{\theta}=0.5$ for Walker2D and $\overline{\theta}=0.1$ Hopper). And $p = 30$ for Walker2D and $p = 50$ for Hopper. When $|\theta| > 2\overline{\theta}$ the robot falls, the episode terminates. The maximum episode length is reduced to 500 steps.
\end{itemize}

The risky robot navigation is proposed by~\cite{ma2021conservative}. The experiments are conducted on a server with two Intel Xeon CPU E5-2697 v2 CPUs with 6 NVIDIA TITAN X Pascal GPUs, 2 NVIDIA TITAN RTX, and 768 GB memory.
\subsection{Risk-Sensitive D4RL}
Firstly, we test the performance of the proposed UDAC in risk-sensitive D4RL environments. Hence, we turn our attention to stochastic D4RL~\cite{urp2021riskaverse}. The original D4RL benchmark~\cite{fu2020d4rl} comprises datasets obtained from SAC agents that exhibit varying degrees of performance (Medium, Expert, and Mixed) on the Hopper, Walker2d, and HalfCheetah in MuJoCo environments~\cite{todorov2012mujoco}. Stochastic D4RL modifies the rewards to reflect the stochastic damage caused to the robot due to unnatural gaits or high velocities; please refer to~\Cref{appendix:expsetup} for details. The Expert dataset consists of rollouts generated by a fixed SAC agent trained to convergence; the Medium dataset is constructed similarly, but the agent is trained to attain only 50\% of the expert agent's return. We have selected several state-of-the-art risk-sensitive offline RL algorithms as well as risk-neutral algorithms. It includes: O-RAAC~\cite{urp2021riskaverse}, CODAC~\cite{ma2021conservative}, O-WCPG~\cite{pmlr-v100-tang20a}, Diffusion-QL~\cite{wang2023diffusion}, BEAR~\cite{kumar2019stabilizing} and CQL~\cite{kumar2020conservative}. Note that, the O-WCPG is not originally designed for offline settings. We follow the same procedure as described in~\cite{urp2021riskaverse} to transform it into an offline algorithm. Among them, ORAAC, CODAC and O-WCPG are risk-sensitive algorithms, Diffusion-QL, BEAR and CQL are risk-natural. Moreover, ORAAC and CODAC are distributional RL algorithms.
% Finally, the Mixed dataset represents the replay buffer of the Medium agent. 

In~\Cref{tab:rs}, we present the mean and $\text{CVaR}_{0.1}$ returns on test episodes, along with the duration for each approach. These results are averaged over 5 random seeds. As shown, UDAC outperforms the baselines on most datasets. It is worth noting that CODAC contains two different variants that optimize different objectives. In this task, we select CODAC-C as the baseline since it targets risk-sensitive situations. Furthermore, we observe that the performance of CQL varies significantly across datasets. Under the risk-sensitive setting, it performs better than two other risk-sensitive algorithms, O-RAAC and CODAC, in terms of $\text{CVaR}_{0.1}$. One possible explanation for this observation is that CQL learns the full distribution, which helps to stabilize the training and thus yields a better $\text{CVaR}_{0.1}$ than O-RAAC and CODAC.
\begin{table*}[ht]
    \centering
     \caption{Performance of the UDAC algorithm on risk-sensitive offline D4RL environments using medium, expert and mixed datasets. We compare the $\text{CVaR}_{0.1}$ and mean of the episode returns, and report the mean (standard deviation) of each metric. Our results indicate that across all environments, UDAC outperforms benchmarks with respect to $\text{CVaR}_{0.1}$. Furthermore, in environments that terminate, UDAC achieves longer episode durations as well.}
     \resizebox{0.82\textwidth}{!}{
    \begin{tabular}{cc|cc|cc|ccccc}
        \hline
        & \multirow{2}{*}{Algorithm} & \multicolumn{2}{c|}{Medium}   & \multicolumn{2}{c|}{Expert} &  \multicolumn{2}{c}{Mixed}  \\ 
         &  & $\text{CVaR}_{0.1} \uparrow$ & Mean $\uparrow$  & $\text{CVaR}_{0.1} \uparrow$ & Mean $\uparrow$& $\text{CVaR}_{0.1} \uparrow$ & Mean $\uparrow$  \\\hline
         \multirow{7}{*}{\rotatebox[origin=c]{90}{Half-Cheetah}}& O-RAAC & \color{blue}{214(36)} & 331(30)  & 595(191) & 1180(78) & 307(6) & 119(27)  \\
         & CODAC & -41(17) & 338(25)  & \color{blue}{687(152)} & \color{blue}{1255(101)} & \color{blue}{396(56)} & \color{blue}{238(59)} \\
         & O-WCPG & 76(14) & 316(23) & 248(232)& 905(107)& 217(33) & 164(76)\\
         & Diffusion-QL & 84(32) & \color{blue}{329(20)}  & 542(102)&  728(104) & 199(43) & 143(77)\\
         & BEAR & 15(30) &  312(20) & 44(20)& 557(15) & 24(30) & 324(59)\\
         & CQL & -15(17) & 33(36)  & -207(47) & -75(23) & 214(52) &  12(24) \\
         & Ours & \color{red}{276(29)} & \color{red}{417(18)}  & \color{red}{732(104)} & \color{red}{1352(100)} & \color{red}{462(48)} & \color{red}{275(50)} \\ \hline
         \multirow{7}{*}{\rotatebox[origin=c]{90}{Walker-2D}}& O-RAAC & 663(124)& 1134(20)& 1172(71)& 2006(56)& 222(37) & -70(76)   \\
         & CODAC &  1159(57) & \color{blue}{1537(78)}& \color{blue}{1298(98)} & \color{blue}{2102(102)}& \color{blue}{450(193)}& \color{blue}{261(231)} \\
         & O-WCPG & -15(41) & 283(37) & 362(33)& 1372(160) & 201(33) & -77(54)\\
         & Diffusion-QL  & 31(10) &  273(20) & 621(49) & 1202(63) & 302(154) & 102(88)\\
         & BEAR & 517(66)& 1318(31)& 1017(49)& 1783(32) & 291(90) & 58(89)\\
         & CQL & \color{blue}{1244(128)}  & 1524(99) & 1301(78) & 2018(65) & 74(77) & -64(-78)\\
         & Ours & \color{red}{1368(89)} & \color{red}{1602(85)} & \color{red}{1398(67)} & \color{red}{2198(78)} & \color{red}{498(91)} & \color{red}{299(194)}\\\hline
         \multirow{7}{*}{\rotatebox[origin=c]{90}{Hopper}}& O-RAAC & \color{blue}{1416(28)} & 1482(4) & 980(28)& 1385(33)& 876(87) & 525(323) \\
         & CODAC & 976(30) & 1014(16)  & 990(19) & 1398(29) & \color{blue}{1551(33)}& \color{blue}{1450(101)} \\
         & O-WCPG & -87(25) &  69(8) &  720(34) & 898(12) & 372(100) & 442(201) \\
         & Diffusion-QL  & 981(28) &  1001(11) & 406(31) & 583(18) & 1021(102) & 1010(59)\\
         & BEAR & 1252(47) & \color{blue}{1575(8)} & 852(30) & 1180(12) & 758(128) & 462(202) \\
         & CQL & 1344(40) & 1524(20) & \color{blue}{1289(30)} & \color{blue}{1402(30)} & 189(63) & -21(62)\\
         & Ours & \color{red}{1502(19)} & \color{red}{1602(18)}  & \color{red}{1302(28)} & \color{red}{1502(27)} & \color{red}{1620(42)} & \color{red}{1523(90)}\\\hline
    \end{tabular}
    }
    \label{tab:rs}
\end{table*}
\subsection{Risky Robot Navigation}
We now aim to test the risk-averse performance of our proposed method. Following the experimental setup detailed in~\cite{ma2021conservative}, we will conduct experiments on two challenging environments, namely Risky Ant and Risky PointMass. In these environments, an Ant robot must navigate from a randomly assigned initial state to the goal state as quickly as possible. However, the path between the initial and goal states includes a risky area, which incurs a high cost upon passing through. While a risk-neutral agent may pass through this area, a risk-sensitive agent should avoid it. For further information on the experimental setup, including details on the environments and datasets, refer to~\ref{appendix:expsetup}. We perform the comparison with the same baselines as the previous section. We present a detailed comparison between CODAC and ORAAC, two state-of-the-art distributional risk-sensitive offline RL algorithms, as well as other risk-neutral algorithms. Our discussion covers a detailed analysis of their respective performance, strengths, and weaknesses.

\begin{figure*}[ht]
    \centering
    \includegraphics[width=\linewidth]{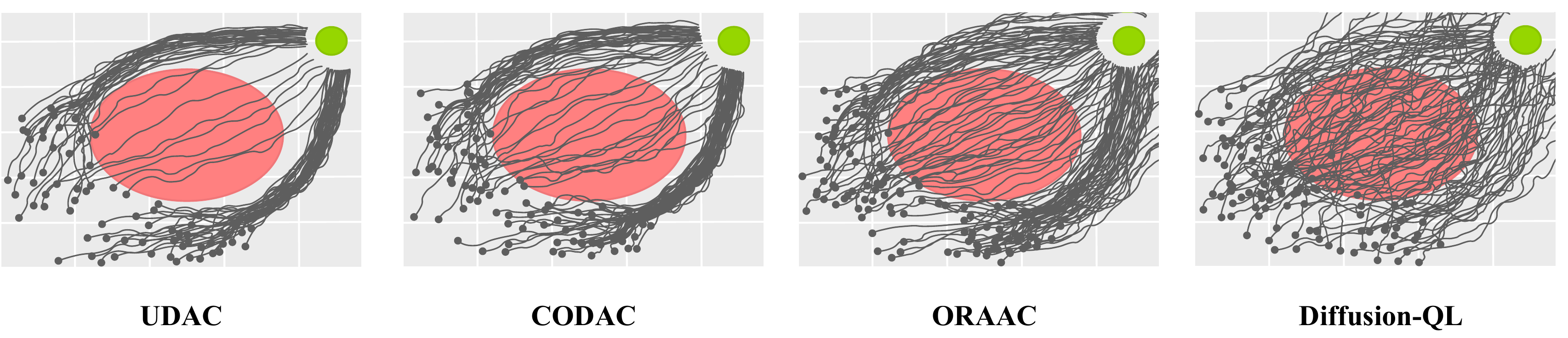}
    \caption{A 2D visualization was created to represent evaluation trajectories on the Risky Ant environment. The risky region is highlighted in red, while the initial states are represented by solid green circles, and the trajectories are shown as grey lines. The results indicate that UDAC consistently approaches the goal while demonstrating the most risk-averse behavior.}
    \label{fig:2driskyant}
\end{figure*}

The effectiveness of each approach is evaluated by running 100 test episodes and measuring various metrics, including the mean, median, and $\text{CVaR}_{0.1}$ returns, as well as the total number of violations (i.e., the time steps spent inside the risky region). To obtain reliable estimates, these metrics are averaged over 5 random seeds. Moreover, we also provide a visualization of the evaluation trajectories on the Risky Ant environment on~\Cref{fig:2driskyant}, and we can see that the CODAC and ORAAC do have some risk-averse action, while UDAC successfully avoids the majority of the risk area. We also note that the Diffusion-QL failed to avoid the risk area as it is not designed for risk-sensitive tasks. 

\begin{table*}[ht]
    \caption{Risky robot navigation quantitative evaluation.}
    \centering
    \resizebox{\textwidth}{!}{
    \begin{tabular}{c|cccc|cccc}
        \hline
        Algorithm & \multicolumn{4}{c|}{Risky PointMass} & \multicolumn{4}{c}{Risky Ant} \\
         & Mean $\uparrow$ & Median $\uparrow$ & $\text{CVaR}_{0.1} \uparrow$ & Violations $\downarrow$ & Mean $\uparrow$& Median $\uparrow$& $\text{CVaR}_{0.1} \uparrow$ & Violations $\downarrow$\\ \hline
         CODAC & \textcolor{blue}{-6.1(1.8)} & -4.9(1.2) & \textcolor{blue}{-14.7(2.4)} & \textcolor{red}{0(0)} & \textcolor{blue}{-456.3(53.2)} & \textcolor{blue}{-433.4(47.2)} & \textcolor{blue}{-686.6(87.2)} & \textcolor{blue}{347.8(42.1)}\\
         ORAAC &  -10.7(1.6) & -4.6(1.3) &  -64.1(2.6) & 138.7(20.2) & -788.1(98.2) & -795.3(86.2) & -1247.2(105.7) & 1196.1(99.6)\\
         CQL & -7.5(2.0) & \textcolor{blue}{-4.4(1.6)} & -43.4(2.8) & \textcolor{blue}{93.4(10.5)} & -967.8(78.2) & -858.5(87.2) & -1887.3(122.4) & 1854.3(130.2)\\
         O-WCPG & -12.4(2.5) & -5.1(1.7) & -67.4(4.0) & 123.5(10.6)  & -819.2(77.4) & -791.5(80.2) & -1424.5(98.2) & 1455.3(104.5)\\ 
         Diffusion-QL & -13.5(2.2) & -5.7(1.3) & -69.2(4.6) & 140.1(10.5) & -892.5(80.3) & -766.5(66.7) & -1884.4(133.2) & 1902.4(127.9) \\
         BEAR & -12.5(1.6) & -5.3(1.2) & -64.3(5.6) & 110.4(9.2) & -823.4(45.2) & -772.5(55.7) & -1424.5(102.3) & 1458.9(100.7) \\
         Ours & \textcolor{red}{-5.8(1.4)} & \textcolor{red}{-4.2(1.3)} & \textcolor{red}{-12.4(2.0)} & \textcolor{red}{0(0)} & \textcolor{red}{-392.5(33.2)} & \textcolor{red}{-399.5(40.5)} & \textcolor{red}{-598.5(50.2)} & \textcolor{red}{291.5(20.8)}\\ \hline
    \end{tabular}
    }
    \label{tab:riskyrobot}
\end{table*}

Our experimental findings indicate that UDAC consistently outperforms other approaches on both the $\text{CVaR}_{0.1}$ return and the number of violations, suggesting that UDAC is capable of avoiding risky actions effectively. Furthermore, UDAC exhibits competitive performance in terms of mean return, attributed to its superior performance in $\text{CVaR}_{0.1}$. A noteworthy observation is that in the Risky PointMass environment, UDAC learns a safe policy that avoids the risky region entirely, resulting in zero violations, even in the absence of such behavior in the training dataset.

\subsection{Risk-Neutral D4RL}
Next, we demonstrate the effectiveness of UDAC even in scenarios where the objective is to optimize the standard expected return. For this purpose, we evaluate UDAC on the widely used D4RL Mujoco benchmark and compare its performance against state-of-the-art algorithms previously benchmarked in \cite{wang2023diffusion}. We include ORAAC and CODAC as offline distributional RL baselines. In our evaluation, we report the performance of each algorithm in terms of various metrics, including the mean and median returns, as well as the $\text{CVaR}_{0.1}$ returns.

% \begin{figure*}[!ht]
%     \centering
%     \begin{subfigure}[b]{0.32\linewidth}
%         \includegraphics[width=\linewidth]{halfcheetah-m.jpg}
%     \end{subfigure}
%     \begin{subfigure}{0.32\linewidth}
%         \includegraphics[width=\linewidth]{hopper-m.jpg}
%     \end{subfigure}
%         \begin{subfigure}{0.32\linewidth}
%         \includegraphics[width=\linewidth]{walker2d-m.jpg}
%     \end{subfigure}
        
%     \begin{subfigure}{0.32\linewidth}
%         \includegraphics[width=\linewidth]{halfcheetah-e.jpg}
%     \end{subfigure}
%     \begin{subfigure}{0.32\linewidth}
%         \includegraphics[width=\linewidth]{hopper-e.jpg}
%     \end{subfigure}
%     \begin{subfigure}{0.32\linewidth}
%         \includegraphics[width=\linewidth]{walker2d-e.jpg}
%     \end{subfigure}
    
%     \caption{The effect of the hyperparameter $\lambda$ on the CVaR of returns varies across risk-sensitive D4RL environments. When $\lambda$ approaches 0, the policy imitates the behavior policy, resulting in poor risk-averse performance. On the other hand, when $\lambda$ approaches 1, the policy suffers from the bootstrapping error, which leads to low performance. The optimal value of $\lambda$ depends on the environment.}
%     \label{fig:hyper}
% \end{figure*}

\begin{figure*}
    \centering
    \includegraphics[width=0.9\linewidth]{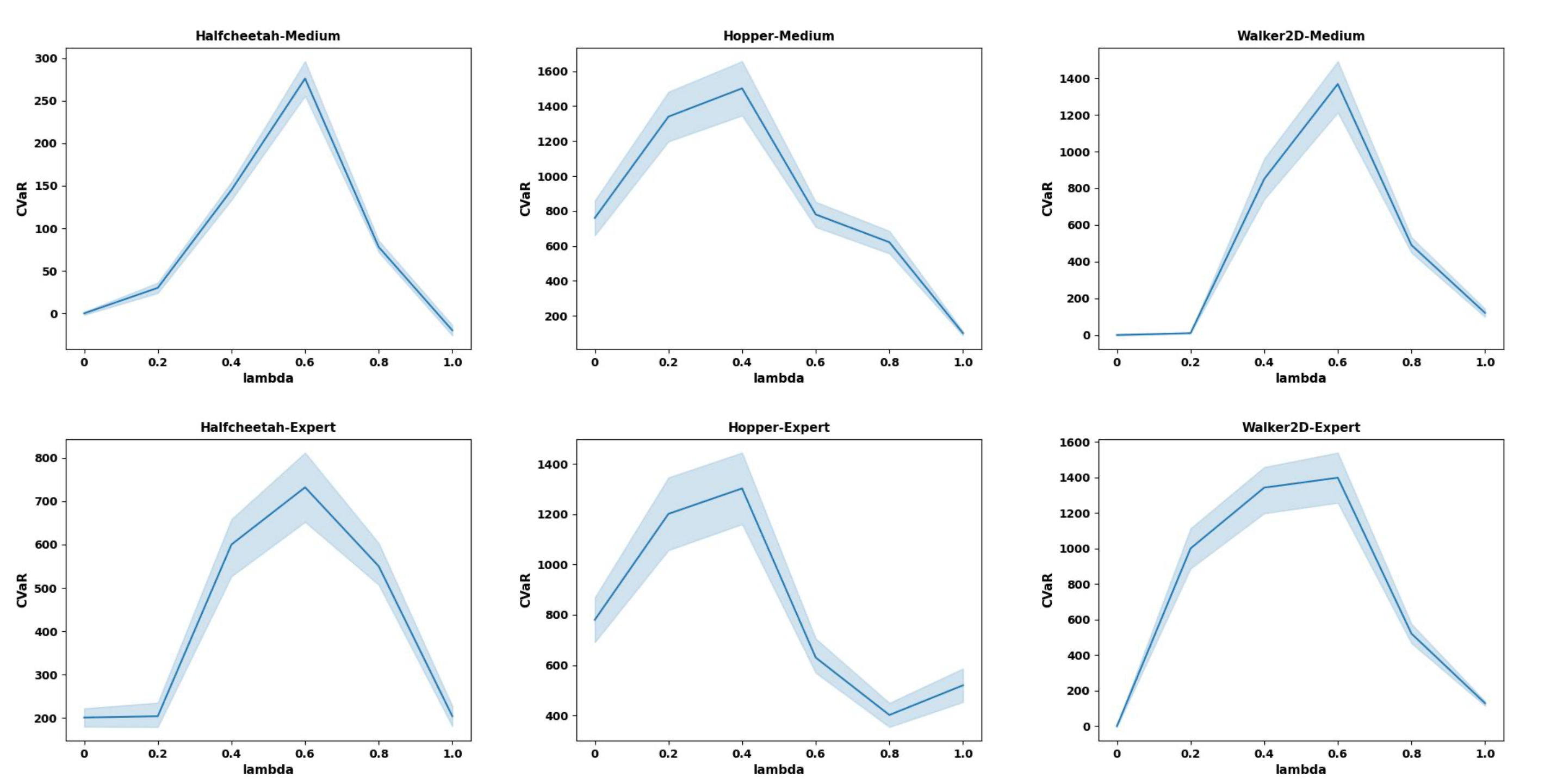}
    \caption{The effect of the hyperparameter $\lambda$ on the CVaR of returns varies across risk-sensitive D4RL environments. When $\lambda$ approaches 0, the policy imitates the behavior policy, resulting in poor risk-averse performance. On the other hand, when $\lambda$ approaches 1, the policy suffers from the bootstrapping error, which leads to low performance. The optimal value of $\lambda$ depends on the environment.}
    \label{fig:hyper}
\end{figure*}

\begin{table*}[ht]
    \centering
        \caption{Normalized Return on the risk-neutral D4RL Mujoco Suite.}
    \resizebox{\textwidth}{!}{
    \begin{tabular}{c|ccccccc}
        \hline
        Dataset & DT & IQL & CQL & Diffusion-QL & ORAAC & CODAC & Ours  \\\hline
         halfcheetah-medium & 42.6(0.3) & 47.4(0.2) & 44.4(0.5) & \textcolor{blue}{51.1(0.3)} & 43.6(0.4) & 48.4(0.4) & \textcolor{red}{51.7(0.2)}\\
         hopper-medium & 67.6(6.2)& 66.3(5.7) & 53.0(28.5) & \textcolor{red}{80.5(4.5)} & 10.4(3.5)  & 72.1(10.2) & \textcolor{blue}{73.2(8.8)} \\
         walker2d-medium & 74.0(9.3)& 78.3(8.7) & 73.3(17.7) & \textcolor{red}{87.0(9.7)} & 27.3(6.2) & 82.0(12.3) & \textcolor{blue}{82.7(10.3)}\\ \hline
         halfcheetah-medium-r & 36.6(2.1) & 44.2(1.2) & 45.5(0.7) & 47.8(1.5)& 38.5(2.0) & \textcolor{blue}{48.6(1.8)} & \textcolor{red}{49.0(1.5)} \\
         hopper-medium-r & 82.7(9.2) & 94.7(8.6) & 88.7(12.9) & \textcolor{red}{101.3(7.7)}& 84.2(10.2) & 95.6(9.7)  & \textcolor{blue}{97.2(8.0)}\\
         walker2d-medium-r & 66.6(4.9) & 73.8(7.1) & 81.8(2.7) & \textcolor{blue}{95.5(5.5)} & 69.2(7.2) & 88.2(9.3) & \textcolor{red}{95.7(6.8)}\\ \hline
         halfcheetah-med-exp & 86.8(4.3) &  86.7(5.3)& 75.6(25.7) & \textcolor{red}{96.8(5.8)} & 24.0(4.3) & 70.4(10.4) & \textcolor{blue}{94.2(8.0)}\\
         hopper-med-exp &107.6(11.7) & 91.5(14.3) & 105.6(12.9) & \textcolor{blue}{111.1(12.6)}& 28.2(4.5)  & 106.0(10.4) & \textcolor{red}{112.2(10.2)}\\
         walker2d-med-exp & 108.1(1.9)& 109.6(1.0)& \textcolor{blue}{111.0(1.6)} & 110.1(2.0)& 18.2(2.0)  &  112.0(6.5) & \textcolor{red}{114.0(2.3)}\\ \hline
    \end{tabular}
    }
    \label{tab:riskneutral}
\end{table*}
As per~\cite{wang2023diffusion}, we directly report results for non-distributional approaches. To evaluate CODAC and ORAAC, we follow the same procedure as~\cite{wang2023diffusion}, training for 1000 epochs (2000 for Gym tasks) with a batch size of 256 and 1000 gradient steps per epoch. We report results averaged over 5 random seeds, as presented in~\Cref{tab:riskneutral}, demonstrate that UDAC achieves remarkable performance across all 9 datasets, outperforming the state-of-the-art on 5 datasets (halfcheetah-medium, halfcheetah-medium-replay, walker2d-medium-replay, hopper-medium-expert, and walker2d-medium-expert). Although UDAC performs similarly to Diffusion-QL in risk-neutral tasks, it is designed primarily for risk-sensitive scenarios, and its main strength lies in such situations. In contrast, UDAC leverages diffusion inside the actor to enable expressive behavior modeling and enhance performance compared to CODAC and ORAAC.

\subsection{Hyperparameter Study in Risk-Sensitive D4RL}
We intend to investigate a crucial hyperparameter within the actor network, namely $\lambda$, in our ongoing research. In~\Cref{fig:hyper}, we present an ablation study on the impact of this hyperparameter in risk-sensitive D4RL. Our findings indicate that selecting an appropriate value of $\lambda$ is critical to achieving high performance as it balances pure imitation from the behavior policy with pure reinforcement learning. Specifically, as $\lambda\rightarrow 0$, the policy becomes more biased towards imitation and yields poor risk-averse performance, while as $\lambda\rightarrow 1$, the policy suffers from bootstrapping errors leading to lower performance.

% \subsection{Different Distorted Operators}\label{sec:distoreted}
% In order to demonstrate the performance of UDAC in different distorted operators, we have selected the risky environment - Risky PointMass and Risky Ant to conduct the experiments. We conduct the experiments using a similar setting as mentioned in previous parts. And the results can be found on~\Cref{sec:appendix-distorted}. We can find that, the UDAC-Wang achieves similar performance with UDAC in the Risky PointMass but slightly worse on Risky Ant. 
% Based on the conclusions of previous studies~\cite{ma2021conservative,dabney2018implicit}, it can be inferred that Wang's risk preferences are marginally more inclined towards risk-seeking when compared to the CVaR approach. This inference can be attributed to the fact that Wang assigns a non-zero weight (although negligible) to quantile values that exceed the risk cutoff threshold. On the other hand, the CPW approach can be regarded as being similar to a risk-neutral strategy. This is because the CPW approach is designed to model human game-play behavior, and thus does not explicitly incorporate risk preferences into its framework.

\section{Conclusion and Future Work}
In this study, we propose a novel model-free offline RL algorithm, namely the Uncertainty-aware offline Distributional Actor-Critic (UDAC). The UDAC framework leverages the diffusion model to enhance its behavior policy modeling capabilities. Our results demonstrate that UDAC outperforms existing risk-averse offline RL methods across various benchmarks, achieving exceptional performance. Moreover, UDAC also achieves comparable performance in risk-natural environments with the recent state-of-the-art methods. Future research directions may include optimizing the sampling speed of the diffusion model~\cite{salimans2022progressive} or other directions related to the optimization of the risk objectives. Moreover, the application of the classifier-free guideline~\cite{ho2022classifier} is worth for investigation.

%%%%%%%%%%%%%%%%%%%%%%%%%%%%%%%%%%%%%%%%%%%%%%%%%%%%%%%%%%%%%%%%%%%%%%%%%%%%%%%%

%%%%%%%%%%%%%%%%%%%%%%%%%%%%%%%%%%%%%%%%%%%%%%%%%%%%%%%%%%%%%%%%%%%%%%%%%%%%%%%%

%%%%%%%%%%%%%%%%%%%%%%%%%%%%%%%%%%%%%%%%%%%%%%%%%%%%%%%%%%%%%%%%%%%%%%%%%%%%%%%%

\bibliographystyle{IEEEtran}
\bibliography{IEEEabrv,sample}

\begin{thebibliography}{10}
\providecommand{\url}[1]{#1}
\csname url@rmstyle\endcsname
\providecommand{\newblock}{\relax}
\providecommand{\bibinfo}[2]{#2}
\providecommand\BIBentrySTDinterwordspacing{\spaceskip=0pt\relax}
\providecommand\BIBentryALTinterwordstretchfactor{4}
\providecommand\BIBentryALTinterwordspacing{\spaceskip=\fontdimen2\font plus
\BIBentryALTinterwordstretchfactor\fontdimen3\font minus \fontdimen4\font\relax}
\providecommand\BIBforeignlanguage[2]{{%
\expandafter\ifx\csname l@#1\endcsname\relax
\typeout{** WARNING: IEEEtran.bst: No hyphenation pattern has been}%
\typeout{** loaded for the language `#1'. Using the pattern for}%
\typeout{** the default language instead.}%
\else
\language=\csname l@#1\endcsname
\fi
#2}}

\bibitem{levine2020offline}
S.~Levine, A.~Kumar, G.~Tucker, and J.~Fu, ``Offline reinforcement learning: Tutorial, review, and perspectives on open problems,'' \emph{arXiv preprint arXiv:2005.01643}, 2020.

\bibitem{urp2021riskaverse}
\BIBentryALTinterwordspacing
N.~A. Urp{\'\i}, S.~Curi, and A.~Krause, ``Risk-averse offline reinforcement learning,'' in \emph{International Conference on Learning Representations}, 2021. [Online]. Available: \url{https://openreview.net/forum?id=TBIzh9b5eaz}
\BIBentrySTDinterwordspacing

\bibitem{lyu2022mildly}
J.~Lyu, X.~Ma, X.~Li, and Z.~Lu, ``Mildly conservative q-learning for offline reinforcement learning,'' in \emph{Thirty-sixth Conference on Neural Information Processing Systems}, 2022.

\bibitem{ma2021conservative}
Y.~Ma, D.~Jayaraman, and O.~Bastani, ``Conservative offline distributional reinforcement learning,'' \emph{Advances in Neural Information Processing Systems}, vol.~34, pp. 19\,235--19\,247, 2021.

\bibitem{pearce2023imitating}
\BIBentryALTinterwordspacing
T.~Pearce, T.~Rashid, A.~Kanervisto, D.~Bignell, M.~Sun, R.~Georgescu, S.~V. Macua, S.~Z. Tan, I.~Momennejad, K.~Hofmann, and S.~Devlin, ``Imitating human behaviour with diffusion models,'' in \emph{The Eleventh International Conference on Learning Representations}, 2023. [Online]. Available: \url{https://openreview.net/forum?id=Pv1GPQzRrC8}
\BIBentrySTDinterwordspacing

\bibitem{wang2023diffusion}
\BIBentryALTinterwordspacing
Z.~Wang, J.~J. Hunt, and M.~Zhou, ``Diffusion policies as an expressive policy class for offline reinforcement learning,'' in \emph{International Conference on Learning Representations}, 2023. [Online]. Available: \url{https://openreview.net/forum?id=AHvFDPi-FA}
\BIBentrySTDinterwordspacing

\bibitem{janner2022planning}
M.~Janner, Y.~Du, J.~B. Tenenbaum, and S.~Levine, ``Planning with diffusion for flexible behavior synthesis,'' \emph{arXiv preprint arXiv:2205.09991}, 2022.

\bibitem{singh2020improving}
R.~Singh, Q.~Zhang, and Y.~Chen, ``Improving robustness via risk averse distributional reinforcement learning,'' in \emph{Learning for Dynamics and Control}.\hskip 1em plus 0.5em minus 0.4em\relax PMLR, 2020, pp. 958--968.

\bibitem{queeney2024risk}
J.~Queeney and M.~Benosman, ``Risk-averse model uncertainty for distributionally robust safe reinforcement learning,'' \emph{Advances in Neural Information Processing Systems}, vol.~36, 2024.

\bibitem{riedmiller2005neural}
M.~Riedmiller, ``Neural fitted q iteration--first experiences with a data efficient neural reinforcement learning method,'' in \emph{Machine Learning: ECML 2005: 16th European Conference on Machine Learning, Porto, Portugal, October 3-7, 2005. Proceedings 16}.\hskip 1em plus 0.5em minus 0.4em\relax Springer, 2005, pp. 317--328.

\bibitem{bellemare2017distributional}
M.~G. Bellemare, W.~Dabney, and R.~Munos, ``A distributional perspective on reinforcement learning,'' in \emph{International conference on machine learning}.\hskip 1em plus 0.5em minus 0.4em\relax PMLR, 2017, pp. 449--458.

\bibitem{dabney2018implicit}
W.~Dabney, G.~Ostrovski, D.~Silver, and R.~Munos, ``Implicit quantile networks for distributional reinforcement learning,'' in \emph{International conference on machine learning}.\hskip 1em plus 0.5em minus 0.4em\relax PMLR, 2018, pp. 1096--1105.

\bibitem{huber1992robust}
P.~J. Huber, ``Robust estimation of a location parameter,'' in \emph{Breakthroughs in statistics: Methodology and distribution}.\hskip 1em plus 0.5em minus 0.4em\relax Springer, 1992, pp. 492--518.

\bibitem{dabney2018distributional}
W.~Dabney, M.~Rowland, M.~Bellemare, and R.~Munos, ``Distributional reinforcement learning with quantile regression,'' in \emph{Proceedings of the AAAI Conference on Artificial Intelligence}, vol.~32, no.~1, 2018.

\bibitem{yang2019fully}
D.~Yang, L.~Zhao, Z.~Lin, T.~Qin, J.~Bian, and T.-Y. Liu, ``Fully parameterized quantile function for distributional reinforcement learning,'' \emph{Advances in neural information processing systems}, vol.~32, 2019.

\bibitem{pratt1978risk}
J.~W. Pratt, ``Risk aversion in the small and in the large,'' in \emph{Uncertainty in economics}.\hskip 1em plus 0.5em minus 0.4em\relax Elsevier, 1978, pp. 59--79.

\bibitem{haarnoja2018soft}
T.~Haarnoja, A.~Zhou, K.~Hartikainen, G.~Tucker, S.~Ha, J.~Tan, V.~Kumar, H.~Zhu, A.~Gupta, P.~Abbeel, \emph{et~al.}, ``Soft actor-critic algorithms and applications,'' \emph{arXiv preprint arXiv:1812.05905}, 2018.

\bibitem{fujimoto2019off}
S.~Fujimoto, D.~Meger, and D.~Precup, ``Off-policy deep reinforcement learning without exploration,'' in \emph{International conference on machine learning}.\hskip 1em plus 0.5em minus 0.4em\relax PMLR, 2019, pp. 2052--2062.

\bibitem{wu2019behavior}
Y.~Wu, G.~Tucker, and O.~Nachum, ``Behavior regularized offline reinforcement learning,'' \emph{arXiv preprint arXiv:1911.11361}, 2019.

\bibitem{pomerleau1991efficient}
D.~A. Pomerleau, ``Efficient training of artificial neural networks for autonomous navigation,'' \emph{Neural computation}, vol.~3, no.~1, pp. 88--97, 1991.

\bibitem{ho2020denoising}
J.~Ho, A.~Jain, and P.~Abbeel, ``Denoising diffusion probabilistic models,'' \emph{Advances in Neural Information Processing Systems}, vol.~33, pp. 6840--6851, 2020.

\bibitem{kong2021fast}
Z.~Kong and W.~Ping, ``On fast sampling of diffusion probabilistic models,'' \emph{arXiv preprint arXiv:2106.00132}, 2021.

\bibitem{song2020score}
Y.~Song, J.~Sohl-Dickstein, D.~P. Kingma, A.~Kumar, S.~Ermon, and B.~Poole, ``Score-based generative modeling through stochastic differential equations,'' \emph{arXiv preprint arXiv:2011.13456}, 2020.

\bibitem{rigter2022one}
M.~Rigter, B.~Lacerda, and N.~Hawes, ``One risk to rule them all: A risk-sensitive perspective on model-based offline reinforcement learning,'' \emph{arXiv preprint arXiv:2212.00124}, 2022.

\bibitem{nichol2021improved}
A.~Q. Nichol and P.~Dhariwal, ``Improved denoising diffusion probabilistic models,'' in \emph{International Conference on Machine Learning}.\hskip 1em plus 0.5em minus 0.4em\relax PMLR, 2021, pp. 8162--8171.

\bibitem{fu2020d4rl}
J.~Fu, A.~Kumar, O.~Nachum, G.~Tucker, and S.~Levine, ``D4rl: Datasets for deep data-driven reinforcement learning,'' \emph{arXiv preprint arXiv:2004.07219}, 2020.

\bibitem{todorov2012mujoco}
E.~Todorov, T.~Erez, and Y.~Tassa, ``Mujoco: A physics engine for model-based control,'' in \emph{2012 IEEE/RSJ International Conference on Intelligent Robots and Systems}.\hskip 1em plus 0.5em minus 0.4em\relax IEEE, 2012, pp. 5026--5033.

\bibitem{pmlr-v100-tang20a}
Y.~C. Tang, J.~Zhang, and R.~Salakhutdinov, ``Worst cases policy gradients,'' in \emph{Proceedings of the Conference on Robot Learning}, ser. Proceedings of Machine Learning Research, L.~P. Kaelbling, D.~Kragic, and K.~Sugiura, Eds., vol. 100.\hskip 1em plus 0.5em minus 0.4em\relax PMLR, 30 Oct--01 Nov 2020, pp. 1078--1093.

\bibitem{kumar2019stabilizing}
A.~Kumar, J.~Fu, M.~Soh, G.~Tucker, and S.~Levine, ``Stabilizing off-policy q-learning via bootstrapping error reduction,'' \emph{Advances in Neural Information Processing Systems}, vol.~32, 2019.

\bibitem{kumar2020conservative}
A.~Kumar, A.~Zhou, G.~Tucker, and S.~Levine, ``Conservative q-learning for offline reinforcement learning,'' \emph{Advances in Neural Information Processing Systems}, vol.~33, pp. 1179--1191, 2020.

\bibitem{salimans2022progressive}
T.~Salimans and J.~Ho, ``Progressive distillation for fast sampling of diffusion models,'' \emph{arXiv preprint arXiv:2202.00512}, 2022.

\bibitem{ho2022classifier}
J.~Ho and T.~Salimans, ``Classifier-free diffusion guidance,'' \emph{arXiv preprint arXiv:2207.12598}, 2022.

\end{thebibliography}

\end{document}